# Determination, Calculation and Representation of the Upper and Lower Sealing Zones During Virtual Stenting of Aneurysms

J. Egger[1,2,3], M. H. A. Bauer[2,3], S. Großkopf [1], C. Biermann[1], B. Freisleben[2], C. Nimsky[3]

[1]Siemens Healthcare, Computed Tomography, Siemensstraße 1, 91301 Forchheim
[2]Dept. of Math. and Computer Science, University of Marburg, Hans-Meerwein-Str., 35032 Marburg
[3]Dept. of Neurosurgery, University of Marburg, Baldingerstrasse, 35033 Marburg

**Purpose**

The minimal-invasive or endovascular treatment of aneurysms requires the selection of an appropriate prosthesis (stent) before an intervention. A stent with wrong dimensions in length or diameter can lead to fatal outcomes for the patient: artery branches can be covered and thus can be excluded from the blood flow. The indicators for a correctly dimensioned stent are the sealing zones where the stent fits to the healthy artery wall. The sealing zones must also have a certain size [1, 2] because an insufficient sealing zone can cause shifting of the stent inside the artery or stress the relevant artery wall. Avoiding stent shifting, stressing of the artery wall and covering branches requires an accurate preoperative stent planning. The importance of this accurate stent planning is indicated by the randomized EndoVascular Aneurysm Repair (EVAR) trials [3]. Within the first four years, 41% of the cases after an endovascular treatment showed complications that required an additional intervention.
In this contribution, a novel method for stent simulation in preoperative computed tomography angiography (CTA) acquisitions of patients is presented where the sealing zones are automatically calculated and visualized. The method is eligible for non-bifurcated and bifurcated stents (Y-stents). Results of the proposed stent simulation with an automatic calculation of the sealing zones for specific diseases (abdominal aortic aneurysms (AAA), thoracic aortic aneurysms (TAA), iliac aneurysms) are presented.
The contribution is organized as follows. Section 2 presents the proposed approach. In Section 3, experimental results are discussed. Section 4 concludes the contribution and outlines areas for future work.

**Methods**

The proposed method starts by computing the vessel centerlines between the user defined seed points [4]: one centerline for a non-bifurcated and two centerlines for a bifurcated stent simulation. To compute the centerlines, a graph-based tracing approach is used, i.e. the image volume is interpreted as a weighted undirected graph. With this representation, vessel tracing becomes a least-cost path problem in graph theory for which well-known algorithms like the Dijkstra-algorithm or the A*-algorithm exist. Based on the computed centerlines, an initial stent is constructed with a geometrical method [5]. For this purpose, rays with a given length are sent out radially from the centerline. The endpoints of the rays are used for the triangulation of the initial stent surface, and the lengths of the rays provide the initial stent diameter. Afterwards, the initial stent is expanded under the influence of several internal and external forces [6]. The formulation of these forces is based on a three-dimensional Active Contour Model (ACM) [7]. The internal forces simulate the elastic behavior of the stent and act in horizontal, vertical and diagonal direction. Contrarily, the external forces pull and push the stent towards the artery wall and consist of: the user defined balloon force, the resistance of the artery wall and the self-collision force for bifurcated stents. Thereby, the recovery of artery wall – lumen and outer wall – is based on a novel minimal closure tracking algorithm that has been validated on over 3000 multiplanar reformatting (MPR) planes in a previous work [8]. After having completely expanded the virtual stent, the sealing zones are calculated automatically with the triangles of the stent model. This is done with the vertices of the triangles that are below a given distance to the artery wall (< 1 mm). Furthermore, the fixed indexing of the surface points of the stent model provides the location of the upper and lower sealing zones. Thus, the different sealing zone areas can automatically be separated, calculated and visualized (Figure 1). For every sealing zone of the horizontal rings of the stent model, the percentage of the fit is provided. Additionally, the total area of every sealing zone (upper / lower) is calculated.

**Results**

To evaluate the proposed method, more than 20 CTA acquisitions from the clinical routine with variations in anatomy and location of pathology – AAA, TAA and iliac aneurysms – as well as phantom data sets were used. Figure 2 shows simulation results for non-bifurcated and bifurcated stents. The sealing zones where the stent fits very tightly to the artery wall are drawn in blue. The non-bifurcated stent from Figure 2a was simulated with a predefined diameter of 19 mm in a CTA dataset with an iliac aneurysm. The whole sealing zone section has an area of 730 $mm^2$. Thus, 200 $mm^2$ belong to the proximal sealing zone and 530 $mm^2$ belong to the distal sealing zone. Figure 2b shows a simulation result where the stent diameter is too small (16 mm). For the chosen diameter, the distal sealing zone is still adequate with an area of 450 $mm^2$. However, in the proximal sealing zone, the stent barely fits to the artery wall (20 $mm^2$). The simulation result shows that a stent with a 3 mm smaller diameter makes an anchorage in the artery impossible, and therefore this stent is not eligible for a subsequent intervention.
Figure 2 (c, d) shows the results of simulated Y-stents for an AAA acquisition. Figure 2c has a sealing zone with an area of 3000 $mm^2$ inside the aorta. The sealing zones in the left and right iliac arteries have areas of 2100 $mm^2$ and 3100 $mm^2$, respectively (bifurcation area: 150 $mm^2$). Figure 2d shows a simulation result where the stent diameter has been chosen too small (< 30 mm for the aorta, < 20 mm for the iliac arteries). Hence, the sealing zones are not adequate after stent expansion (< 300 $mm^2$), and the stent is not applicable for an intervention.

**Conclusion**

The presented method is useful for the preoperative simulation of different stents in CTA datasets to determine whether the stent has adequate dimensions for an intervention. The sealing zones are very important because the stent anchors in these parts of the artery. The automatic visualization additionally shows whether the stent only fits to one side of the artery wall during expansion. The introduced method has been realized as a clinical prototype in C++ within the MeVisLab platform that is currently being used in a hospital for an evaluation under clinical conditions.

There are several areas of future work. For example, the method can be enhanced for patient data where the aneurysm neck – below the renal branches and above the aneurysm sack – is too short for anchoring the stent in the upper sealing zone. Here, the stent is planned over the renal branches with fenestration [9]. Another application area for the presented method are abnormal narrowings in blood vessels (stenosis) [10]. In addition to the sealing zones, the widened area of the artery – in the stenosis section – can be automatically calculated and visualized.

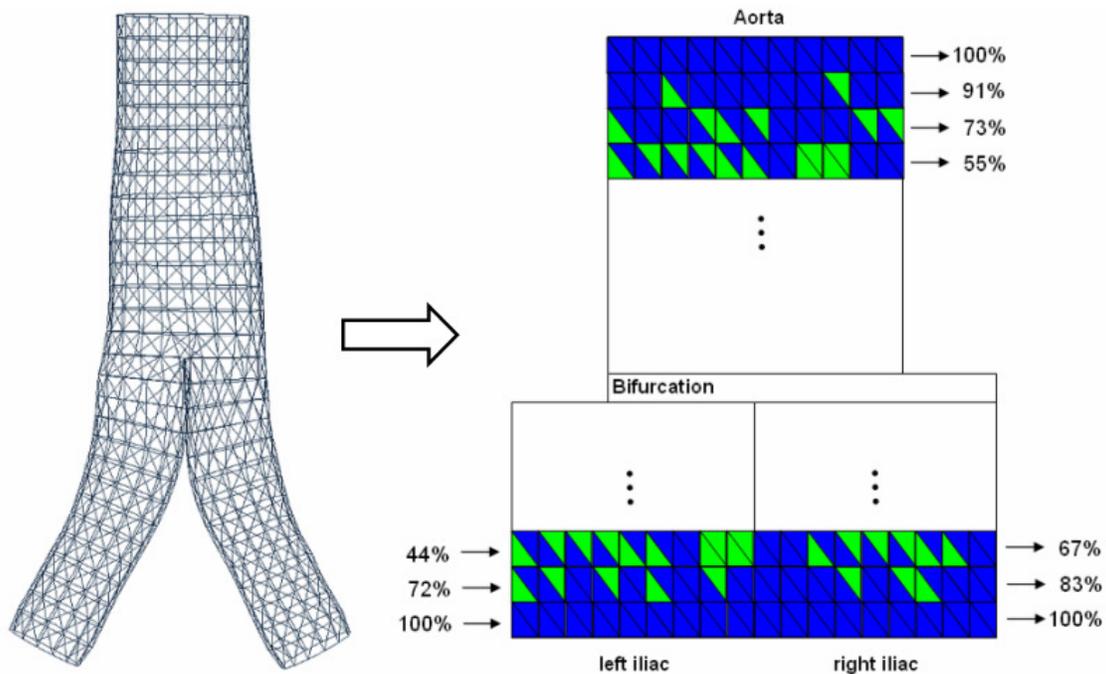

**Figure 1** – Bifurcated stent model (left). Unfolded stent surface (right) with sealing zones drawn in blue (tight fit). Green implies a loose fit (distance to the artery wall > 1 mm).

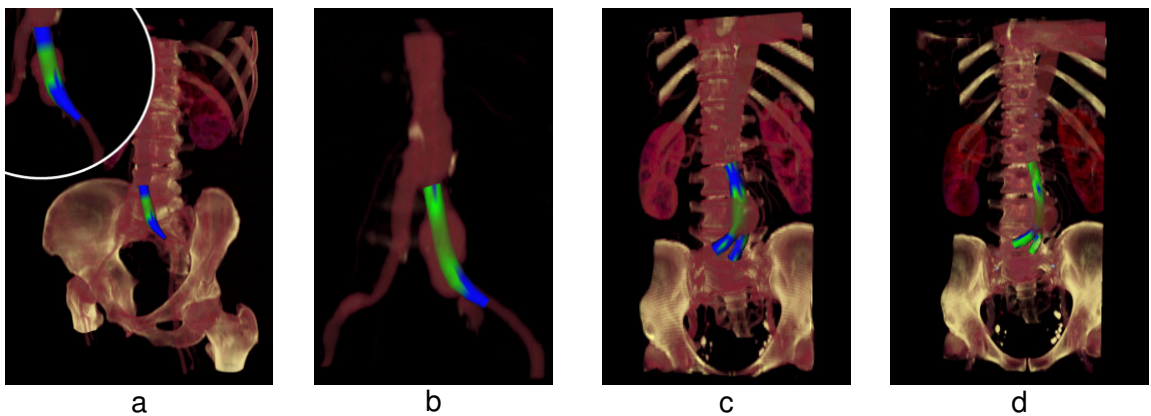

a  b  c  d

**Figure 2** – Simulation results of I-stents (a, b) and Y-stents (c, d). The color coding indicates the goodness of fit of the simulated stent model to the artery wall. Green implies a loose fit, blue a tight fit.